\documentclass[journal,transmag]{IEEEtran}
\ifCLASSINFOpdf
  \usepackage[pdftex]{graphicx}
\else
\fi
\usepackage{amsmath}
 \usepackage[caption=false,font=normalsize,labelfont=sf,textfont=sf]{subfig}
\begin{document}
%
\title{Diffeomorphic Particle Image Velocimetry}



\author{\IEEEauthorblockN{Yong Lee\IEEEauthorrefmark{1}, and
Shuang Mei\IEEEauthorrefmark{2}}
\IEEEauthorblockA{\IEEEauthorrefmark{1}School of Computing, National University of Singapore, NUS, Singapore}
\IEEEauthorblockA{\IEEEauthorrefmark{2}School of Mechanical Engineering and Electronic Information, China University of Geosciences, Wuhan, P.R.China}
\thanks{Manuscript received December X, 2021; revised August X, 2021. 
Corresponding author: S. Mei (email: meishuang@cug.edu.cn).}}

\markboth{IEEE TRANSACTIONS ON INSTRUMENTATION AND MEASUREMENT,~Vol.~xx, August~2021}%
{Lee \MakeLowercase{\textit{et al.}}: Diffeomorphic Particle Image Velocimetry}
%



\IEEEtitleabstractindextext{%
\begin{abstract}


The existing particle image velocimetry (PIV) techniques do not consider the curvature effect of the non-straight particle trajectory, because it seems to be impossible to obtain the curvature information from a pair of particle images. As a result, the computed vector underestimates the real velocity due to the straight-line approximation, that further causes a systematic error for the PIV instrument.
In this work, the particle curved trajectory between two recordings is firstly explained with the streamline segment of a steady flow (diffeomorphic transformation) instead of a single vector, and this novel idea is termed as \textit{diffeomorphic PIV}. Specifically, a deformation field is introduced to describe the particle displacement along the streamline, i.e., we try to find the optimal velocity field, of which the corresponding deformation vector field agrees with the particle displacement.  
Because the variation of the deformation function can be approximated with the variation of the velocity function, the diffeomorphic PIV can be implemented as special iterative PIV. That says, the diffeomorphic PIV warps the images with deformation vector field instead of velocity field, and keeps the rest procedures as same as a conventional iterative PIV. Similar to forward difference interrogation (FDI) and central difference interrogation (CDI), two diffeomorphic deformation schemes ---forward diffeomorphic deformation interrogation (FDDI) and central diffeomorphic deformation interrogation (CDDI)---  are proposed in this paper.
Tested on synthetic images of Lamb-Oseen flows and sine flows, the FDDI achieves significant accuracy improvement across different one-pass displacement estimators (cross-correlation, optical flow, deep learning flow). Besides, the results on three real PIV image pairs demonstrate the non-negligible curvature effect for CDI-based measurement, and our FDDI provides larger velocity estimation---more accurate--- in the fast curvy streamline areas.
The significant accuracy improvement of the combination of FDDI and accurate dense estimator (e.g., optical flow) means that our diffeomorphic PIV paves a completely new way for complex flow field measurement.

\end{abstract}

}

\maketitle

\IEEEdisplaynontitleabstractindextext

%
\IEEEpeerreviewmaketitle

\section{Introduction}
\label{sec_1}
\IEEEPARstart{P}{article} image velocimetry (PIV) is an important experimental fluid instrument that outputs a velocity field, by measuring the image displacement of particles in a predefined time interval $\Delta t$~\cite{adrian1984scattering,raffel2018particle}.
To gain more accurate flow information about the fluids, both the dynamic spatial range (DSR) and dynamic velocity range (DVR) of PIV measurement are encouraged to be maximized. A set of algorithms are thus introduced, including single-pixel ensemble correlation~\cite{billy2004single,westerweel2004single}, optical flows~\cite{quenot1998particle,corpetti2006fluid,lu2021accurate}, deep learning flows~\cite{lee2017piv,cai2019dense,zhang2020unsupervised,stulov2021neural}.  These methods could assign each pixel a vector and thus yield a very large DSR~\cite{kahler2012resolution}. 
The large DVR is often achieved by increasing particle shift---setting a large time interval $\Delta t$, because it is is difficult to further improve the accuracy of PIV estimation beyond $0.02$ pixel (RMSE). 
However, a large time interval $\Delta t$ could cause a non-negligible systematic error (bias) due to the effect of streamline curvature~\cite{kahler2012resolution,scharnowski2013effect,raffel2018piv}. 

\begin{figure}
  \includegraphics{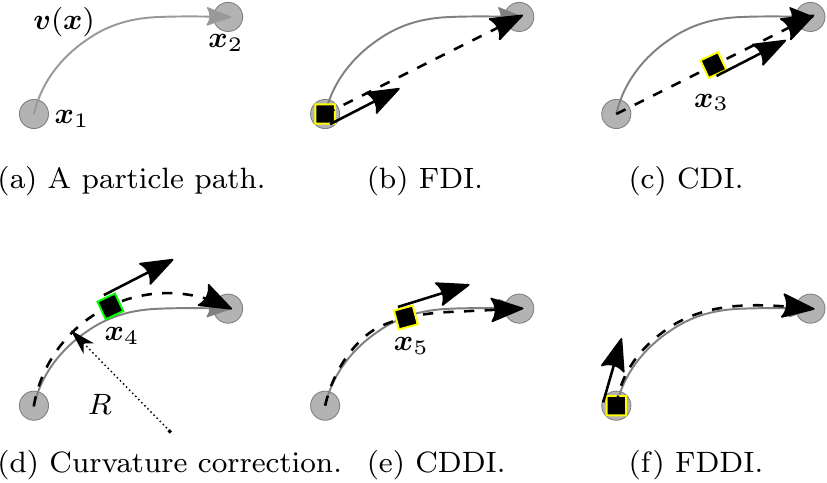}
\caption{The curved particle path (a) and different approximations (dash lines). The FDI (b) and CDI (c) \cite{wereley2001second} treat the path as a straight-line. The curvature correction method~\cite{scharnowski2013effect} approximates the path with a circular arc (d). Our CDDI (e) and FDDI (f) use the streamline of a steady flow to infer the path}
\label{fig_1}       
\end{figure}

As illustrated in Fig.~\ref{fig_1}(a), the effect of streamline curvature~\cite{kahler2012resolution} causes a significant systematic error, because the assumed straight-line estimation between two recording positions $(\boldsymbol{x}_1, \boldsymbol{x}_2)$ does not well describe the true curved particle path. 
However, the curved trajectories certainly happen in the case of non-uniform flow, and the error will obviously increase if the particle travels a longer distance. 
In this paper, we thus focus on the challenging
 systematic error caused by this curvature effect. 

A solid body rotation model of cylindrical Couette flow is employed to quantify the curvature effect~\cite{wereley2001second}. This model points out that the error of forward difference interrogation (FDI) method is proportional to $\Delta t$ while that of central difference interrogation (CDI) is proportional to $(\Delta t)^2$~\cite{wereley2001second}. As shown in Fig.~\ref{fig_1}(b) and (c), both FDI and CDI assume a straight-line trajectory, and the distinct characteristic of CDI is to change the velocity location from $\boldsymbol{x}_1$ to $\boldsymbol{x}_3$.
The solid body rotation model also bridges the true cylindrical Couette flow velocity and the corresponding measurement of CDI. 
Therefore, the curvature correction is possible by inferring the curvature $(1/R)$~\cite{scharnowski2013effect}. As shown in Fig.~\ref{fig_1}(d), the curvature correction assumes the particle follows the circular arc path of an inferred curvature $(1/R)$. 
It is obvious that neither the straight line nor the circular arc could accurately describe the real complex particle trajectory. 

\begin{figure}
  \includegraphics{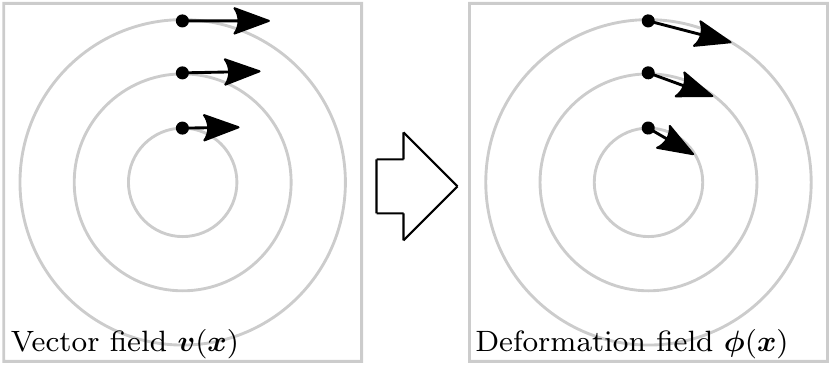}
\caption{The difference between velocity vector field and deformation field for an illustrative Lamb-Oseen flow (circular streamlines). The instantaneous velocity determines the cumulative deformation}
\label{fig_2}       
\end{figure}

Diffeomorphic registration~\cite{joshi2000landmark,beg2005computing,arsigny2006log,balakrishnan2019voxelmorph} provides an invertible geometrical transformation (diffeomorphism), which adopts a steady velocity field to explain the image deformation as a time-dependent process. That says, the diffeomorphic deformation field---deduced from an instantaneous velocity field--- is responsible for the image deformation, as shown in Fig.~\ref{fig_2}. Diffeomorphic registrations achieve competitive performance over large curved deformation and provide physically admissible results~\cite{joshi2000landmark,beg2005computing}. The interesting diffeomorphic deformation will be detailed in section~\ref{sec_2}. Our insight is that the diffeomorphic deformation considers the curved particle trajectory, i.e., the curved particle trajectory could be accurately approximated by the streamline of a velocity field. Thus, the particle displacement can be modeled as integration along the streamline, as shown in Fig.~\ref{fig_1}(e) and (f). As far as we known, neither the diffeomorphic registration nor diffeomorphic deformation has been considered in existing PIV techniques. 

In this work, the \textit{diffeomorphic PIV} is proposed to combine the diffeomorphic deformation and PIV technique. Our diffeomorphic PIV employs a (high DSR) velocity field---instead of a single velocity vector---to explain the particle displacement. Specifically, two diffeomorphic deformation schemes---forward diffeomorphic deformation interrogation (FDDI) and central diffeomorphic deformation interrogation (CDDI)--- are implemented. The main contributions are:
%
\begin{itemize}
  \item The difference between the velocity field and deformation field is clarified for PIV community. 
  \item The diffeomorphic PIV is proposed to estimate the exact velocity field instead of conventional deformation field.
  \item Two image deformation schemes (FDDI and CDDI) for the diffeomorphic PIV are implemented. On various test cases, the FDDI achieves significant accuracy improvement across different one-pass displacement estimators. 
\end{itemize}
The rest of paper is arranged as follows. Section~\ref{sec_2} is devoted to the related works. Section~\ref{sec_3} details the diffeomorphic PIV technique. Section~\ref{sec_4} demonstrates the promising experimental results, followed by some concluding remarks in Section~\ref{sec_5}.

\section{Related works}
\label{sec_2}
\subsection{The velocity field and deformation field}
Given a particle path $\boldsymbol{x}(t), t\in [0,1]$, the velocity ($\boldsymbol{v}(t)=\frac{\partial \boldsymbol{x}(t)}{\partial t}$) is an instantaneous quantity while the displacement $(\boldsymbol{x}(1)-\boldsymbol{x}(0))$ is an integral quantity. So are the velocity vector field $\boldsymbol{v}(\boldsymbol{x})$ and deformation field $\boldsymbol{\phi}(\boldsymbol{x})$, as shown in Fig.~\ref{fig_2}. 
The difference between them has been noted 20 years ago~\cite{wereley2001second}, and the pioneers~\cite{wereley2001second,scharnowski2013effect} have proved that the deformation field computed with CDI scheme approximates the velocity vector field well. Since then, the velocity vector field and deformation field have seldom been distinguished~\cite{lee2017piv,scharnowski2020particle,lu2021accurate}.  Under the circumstance of high DSR techniques, we argue that clear clarifying the relationship of velocity field and the deformation field is beneficial to achieve an accurate PIV measurement.



Actually, it is not a challenging problem to obtain the deformation field by integrating a velocity field along the streamlines.
For the computational fluid dynamics (CFD) community, it is known as Lagrangian particle tracking~(LPT) \cite{guo2004simulation} or discrete particle simulation~(DPS) \cite{tsuji1993discrete}. 
For the image processing community, it is called as diffeomorphic deformation which has been widely used for medical image registration.

\subsection{Diffeomorphic deformation}
Let $\boldsymbol{v}^{(t)}(\boldsymbol{x})$ be the time-dependent velocity field, the particles' position $\boldsymbol{\psi}^{(t)}(\boldsymbol{x})$ is defined through an ordinary differential equation (ODE), or transport equation~\cite{joshi2000landmark,beg2005computing}:
\begin{equation}
\frac{\partial \boldsymbol{\psi}^{(t)}}{\partial t} = \boldsymbol{v}^{(t)}(\boldsymbol{\psi}^{(t)})
\end{equation}
where $t\in [0, 1]$ denotes the time between two recordings. The particles do not change their positions at $t=0$, i.e., $\boldsymbol{\psi}^{(0)}(\boldsymbol{x})$ $=\boldsymbol{x}$. At $t=1$, the particles move to the position $\boldsymbol{\psi}^{(1)}(\boldsymbol{x})$, following the path line of velocity field $\boldsymbol{v}^{(t)}(\boldsymbol{x})$. Provided the particle positions, the deformation field is thus defined as 
\begin{equation}
\begin{split}
\boldsymbol{\phi}_{\boldsymbol{v}}(\boldsymbol{x}) &= \boldsymbol{\psi}^{(1)}(\boldsymbol{x})-\boldsymbol{\psi}^{(0)}(\boldsymbol{x})\\
&= \boldsymbol{\psi}^{(1)}(\boldsymbol{x})-\boldsymbol{x}\\
\end{split}
\end{equation}
where the subscript $\boldsymbol{v}$ of in $\boldsymbol{\phi}_{\boldsymbol{v}}$ is used to explicitly denote the dependence of $\boldsymbol{\phi}$ on the velocity vector field $\boldsymbol{v}$.

During a small time interval, it is reasonable to assume a steady flow, i.e., $\boldsymbol{v}^{(t)}(\boldsymbol{x})=\boldsymbol{v}(\boldsymbol{x})$. Hence, the problem becomes,
\begin{equation}
\frac{\partial \boldsymbol{\psi}^{(t)}}{\partial t} = \boldsymbol{v}(\boldsymbol{\psi}^{(t)})
\end{equation}
Given the $\boldsymbol{v}(\boldsymbol{x})$ and initial condition $\boldsymbol{\psi}^{(0)}(\boldsymbol{x})$ $=\boldsymbol{x}$, there are several methods to solve this ODE.  The \textit{Scaling and Squaring} approach~\cite{arsigny2006log}, a fast integration of stationary ODEs on regular grids, is adopted to estimate the integral quantity $\boldsymbol{\psi}^{(1)}(\boldsymbol{x})$ and $\boldsymbol{\phi}_{\boldsymbol{v}}(\boldsymbol{x})$ in this work. 



\subsection{Iterative PIV}
The general PIV estimation is to find the optimal displacement $\boldsymbol{v}$ that minimizes a distance metric $d(\cdot,\cdot)$ between two warped images. Mathematically, the widely used FDI and CDI implementations could be formulated as:
\begin{equation}
\label{eq_4}
\begin{split}
\boldsymbol{v} &= \mathop{\arg\min}_{\hat{\boldsymbol{v}}} d(\boldsymbol{I}_1(\boldsymbol{x}), \boldsymbol{I}_2(\boldsymbol{x}+ \hat{\boldsymbol{v}}))\textrm{, FDI}\\
\boldsymbol{v} &= \mathop{\arg\min}_{\hat{\boldsymbol{v}}} d(\boldsymbol{I}_1(\boldsymbol{x}- \frac{1}{2}\hat{\boldsymbol{v}}), \boldsymbol{I}_2(\boldsymbol{x}+ \frac{1}{2}\hat{\boldsymbol{v}}))\textrm{, CDI}\\
\end{split}
\end{equation}
where $\boldsymbol{I}_1, \boldsymbol{I}_2$ are two paired recordings. And different distance metric $d(\cdot, \cdot)$ corresponds to different flow estimator, and it is not the focus of the present work. 
An iterative refinement becomes a common trick to obtain an accurate solution. That says, 
\begin{equation}
\boldsymbol{v}_{k+1} = \boldsymbol{v}_{k} + \Delta \boldsymbol{v} 
\end{equation}
with 
\begin{equation}
\begin{split}
\Delta \boldsymbol{v} &= \mathop{\arg\min}_{\Delta \hat{\boldsymbol{v}}} d(\boldsymbol{I}_1(\boldsymbol{x}), \boldsymbol{I}_2(\boldsymbol{x}+ \boldsymbol{v}_k+ \Delta \hat{\boldsymbol{v}})) \\
&= \mathop{\arg\min}_{\Delta \hat{\boldsymbol{v}}} d(\boldsymbol{I}_1(\boldsymbol{x}), \boldsymbol{I}_2^k(\boldsymbol{x}+ \Delta \hat{\boldsymbol{v}}))\textrm{, FDI}\\
\Delta \boldsymbol{v} &= \mathop{\arg\min}_{\Delta \hat{\boldsymbol{v}}} d(\boldsymbol{I}_1(\boldsymbol{x}- \frac{1}{2}(\boldsymbol{v}_k+ \Delta \hat{\boldsymbol{v}})), \\
&\quad\quad\quad\quad\quad\quad \boldsymbol{I}_2(\boldsymbol{x}+ \frac{1}{2}(\boldsymbol{v}_k+ \Delta \hat{\boldsymbol{v}})))\\
&= \mathop{\arg\min}_{\Delta \hat{\boldsymbol{v}}} d(\boldsymbol{I}_1^k(\boldsymbol{x}- \frac{1}{2} \Delta \hat{\boldsymbol{v}}),  \boldsymbol{I}_2^k(\boldsymbol{x}+ \frac{1}{2} \Delta \hat{\boldsymbol{v}}))\textrm{, CDI}\\
\end{split}
\end{equation}
where $\boldsymbol{I}_1^k, \boldsymbol{I}_2^k$ are the warped images with previous vector field $\boldsymbol{v}_k$.
Due to the fast convergence and attractive accuracy, the iterative PIVs have achieved general recognition~\cite{scarano2001iterative,wereley2001second}.




\section{Diffeomorphic PIV}
\label{sec_3}

\subsection{Problem formulation}
As mentioned in Section~\ref{sec_1}, the conventional iterative PIV methods do not tell the velocity field and deformation field apart. 
We thus introduce the \textit{diffeomorphic PIV} which separates the velocity field and deformation field. 
Similar to FDI and CDI, two instances of diffeomorphic PIV (FDDI and CDDI) are formulated with corresponding objectives,   
\begin{equation}
\label{eq_7}
\begin{split}
\boldsymbol{v} &= \mathop{\arg\min}_{\hat{\boldsymbol{v}}} d(\boldsymbol{I}_1(\boldsymbol{x}), \boldsymbol{I}_2(\boldsymbol{x}+  \boldsymbol{\phi_{\hat{\boldsymbol{v}}}})) \textrm{, FDDI}\\
\boldsymbol{v} &= \mathop{\arg\min}_{\hat{\boldsymbol{v}}} d(\boldsymbol{I}_1(\boldsymbol{x}+  \boldsymbol{\phi}_{-0.5\hat{\boldsymbol{v}}}), \boldsymbol{I}_2(\boldsymbol{x}+  \boldsymbol{\phi}_{0.5\hat{\boldsymbol{v}}})) \textrm{, CDDI}\\
\end{split}
\end{equation}
Compared with Eq.~(\ref{eq_4}), the particle displacement is described with the deformation field $\boldsymbol{\phi}(\boldsymbol{x})$ in diffeomorphic PIV.
The deformation field approximates the curved particle trajectory with the streamline of a velocity field.
And the difference between FDDI and CDDI objectives results in different estimated positions, as demonstrated in Fig.~\ref{fig_1}. 
The solution to these objectives---diffeomorphic registration~\cite{joshi2000landmark,beg2005computing}---depends on the specified distance metric $d(\cdot,\cdot)$. 

\subsection{Iterative optimization}

\begin{figure}
\centering
  \includegraphics[width=0.75\columnwidth]{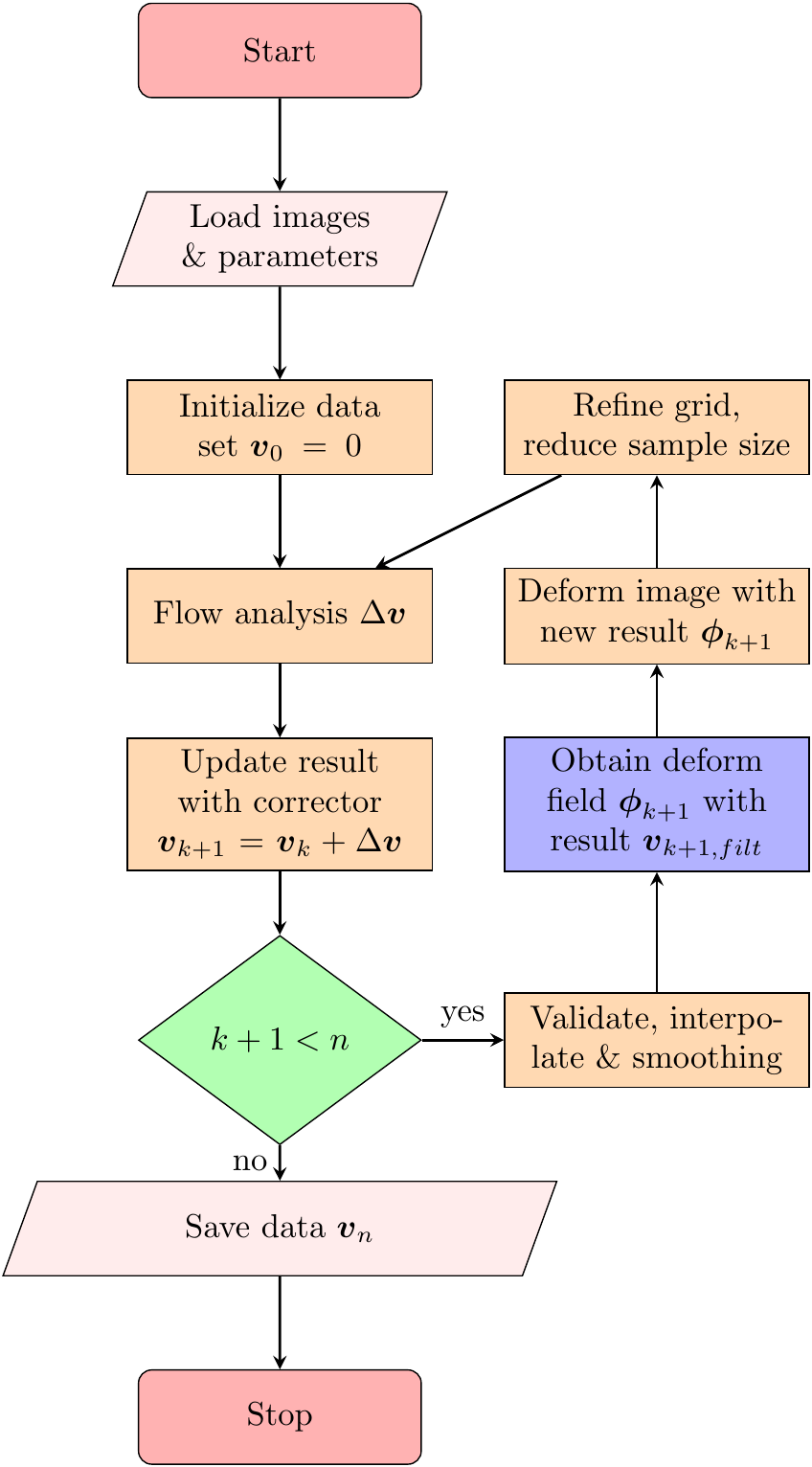}
\caption{Block diagram of our diffeomorphic PIV. It is slightly modified from the conventional iterative PIV~\cite{raffel2018particle}}
\label{fig_3}       
\end{figure}

\begin{table}
\caption{The warped images for different PIV techniques}
\label{tab_0}       
\centering
\begin{tabular}{|c|l|l|}
\hline
Method & $\boldsymbol{I}_1^k(\boldsymbol{x})$ &  $\boldsymbol{I}_2^k(\boldsymbol{x})$ \\
\hline
FDI  &  $\boldsymbol{I}_1(\boldsymbol{x})$  & $\boldsymbol{I}_2(\boldsymbol{x}+\boldsymbol{v}_k)$    \\
\hline
CDI  & $\boldsymbol{I}_1(\boldsymbol{x}-0.5\boldsymbol{v}_k)$  & $\boldsymbol{I}_2(\boldsymbol{x}+0.5\boldsymbol{v}_k)$    \\
\hline
CDDI  & $\boldsymbol{I}_1(\boldsymbol{x}+\boldsymbol{\phi}_{-0.5\boldsymbol{v}_k)}$  & $\boldsymbol{I}_2(\boldsymbol{x}+ \boldsymbol{\phi}_{0.5\boldsymbol{v}_k})$    \\
\hline
FDDI  & $\boldsymbol{I}_1(\boldsymbol{x})$ & $\boldsymbol{I}_2(\boldsymbol{x}+ \boldsymbol{\phi}_{\boldsymbol{v}_k})$    \\ 
\hline
\end{tabular}
\end{table}

To find a general solution to the problems (Eq.~(\ref{eq_7})), the iterative refinement trick is also considered here. 
For the simplicity, we only detail the solution to the FDDI objective. 
Similar to the iterative PIV (as detailed in Section~\ref{sec_2}), an iterative refinement is considered. 
\begin{equation}
\begin{split}
\boldsymbol{v}_{k+1} &= \boldsymbol{v}_{k} + \Delta \boldsymbol{v} \\
\Delta \boldsymbol{v} &= \mathop{\arg\min}_{\Delta \hat{\boldsymbol{v}}} d(\boldsymbol{I}_1(\boldsymbol{x}), \boldsymbol{I}_2(\boldsymbol{x}+  \boldsymbol{\phi_{\boldsymbol{v}_k + \Delta \hat{\boldsymbol{v}}}}))\\
\end{split}
\end{equation}
Because the variation of the deformation function can be approximated with the variation of the velocity function, i.e., $\Delta \boldsymbol{\phi} = \boldsymbol{\phi_{\boldsymbol{v}_k+\Delta \hat{\boldsymbol{v}}}} - \boldsymbol{\phi_{\boldsymbol{v}_k}\approx\Delta \hat{\boldsymbol{v}}}$, 
the velocity corrector finally becomes,  
\begin{equation}
\begin{split}
\Delta \boldsymbol{v} &= \mathop{\arg\min}_{\Delta \hat{\boldsymbol{v}}} d(\boldsymbol{I}_1(\boldsymbol{x}), \boldsymbol{I}_2(\boldsymbol{x}+  \boldsymbol{\phi_{\boldsymbol{v}_k} + \Delta \hat{\boldsymbol{v}}}))\\
&= \mathop{\arg\min}_{\Delta \hat{\boldsymbol{v}}} d(\boldsymbol{I}_1(\boldsymbol{x}), \boldsymbol{I}_2^k(\boldsymbol{x}+ \Delta \hat{\boldsymbol{v}}))\\
\end{split}
\end{equation}
where $\boldsymbol{I}_2^k(\boldsymbol{x}) = \boldsymbol{I}_2(\boldsymbol{x}+  \boldsymbol{\phi}_{\boldsymbol{v}_k})$ is the warped image with previous deformation field $\boldsymbol{\phi}_{\boldsymbol{v}_k}$. 
%
Since this iterative refinement works over different distance metrics,
various flow estimators (CC, OF, DP) can be adopted to estimate the corrector $\Delta \boldsymbol{v}$ with warped images $\boldsymbol{I}_1(\boldsymbol{x})$ and $\boldsymbol{I}^k_2(\boldsymbol{x})$. 
Table.~\ref{tab_0} lists different warping schemes for FDI, CDI, CDDI, and FDDI. 
%
It means that only a slight change is needed to perform diffeomorphic PIV, for a conventional iterative PIV software.
As shown in Fig.~\ref{fig_3}, our diffeomorphic PIV and iterative PIVs could share the block diagram, except the step to generate warped images (block with light blue background).  

\subsection{Error analysis}

\begin{figure}
  \includegraphics{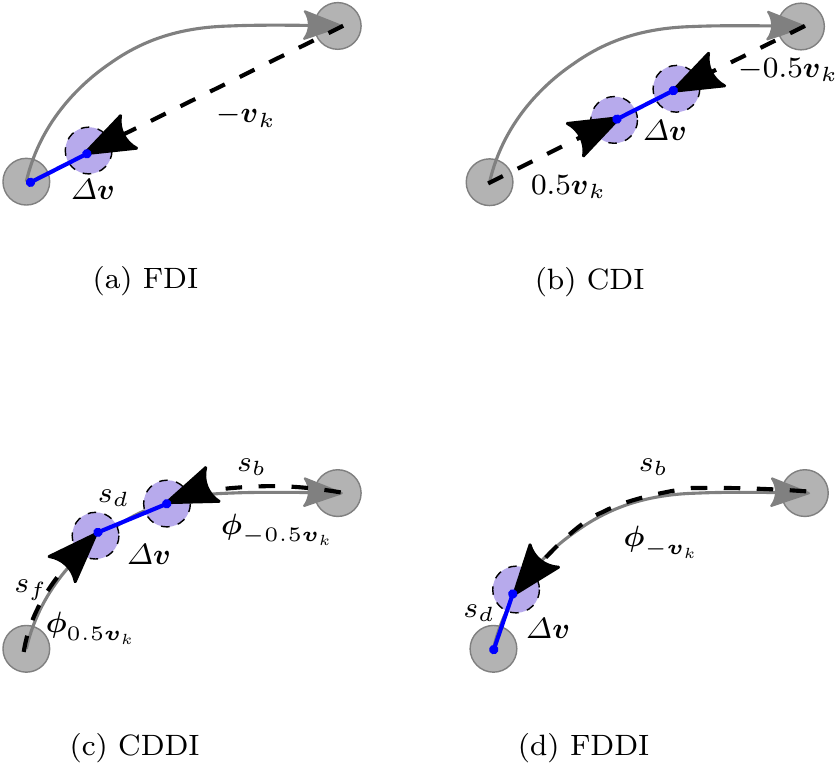}
\caption{The iterative procedures for different image deformation schemes. The dash circles filled with light blue denote the warped particle images, which follow the dashed arrow path
}
\label{fig_4}       
\end{figure}

Because the error analysis helps to understand the performance of FDDI and CDDI, the velocity updates are thus investigated, as shown in Fig.~\ref{fig_4}. In the view of particle path, the length of a curved path  is a summation of several segments.


\begin{equation}
\begin{split}
s_{FDDI} &= s_b+s_d\\
s_{CDDI} &= s_b + s_f+s_d
\end{split}
\end{equation}
where $s_d=|\delta \boldsymbol{v}|$ is the estimated displacement amplitude between the two warped images,  the $s_b$ and $s_f$ denote the backward and forward lengths along the streamline of velocity vector field $\boldsymbol{v}_k$. Hence, the cumulative error $\delta$ depends on the accuracy of each component segment, i.e., 
\begin{equation}
\begin{split}
\delta_{FDDI} &= s-s_{FDDI}
 =\delta s_b+\delta s_d\\
\delta_{CDDI} &= s-s_{CDDI}
  =\delta s_b + \delta s_f+ \delta s_d
\end{split}
\end{equation}
where $s$ is the unknown truth of length. 
It is clear that the error depends on image warping accuracy $(\delta s_b, \delta s_f)$ and the accuracy of displacement estimator $(\delta_{algo}=\delta s_d)$. 
According to the iterative implementation, the image warping error is related to deformation field generation $\delta_{field}$ and the warped image interpolation $\delta_{interp}$. 
Hence,
\begin{equation}
\label{eq_10}
\begin{split}
\delta_{FDDI} &= \delta_{field} + \delta_{interp} + \delta_{algo}\\
\delta_{CDDI} &= 2\delta_{field} + 2\delta_{interp} + \delta_{algo}\\
\end{split}
\end{equation}  
Therefore, an accurate dense estimator is recommended due to the small $\delta_{field}$ and  small $\delta_{algo}$. That says, the dense vector field---computed by high DSR algorithm---enables a more accurate deformation field than that of sparse velocity field. Note that both the algorithm's accuracy and the spatial resolution have effect on the final performance. From the Eq.~(\ref{eq_10}), the FDDI is expected to be more accurate than CDDI due to fewer error sources. A full study on the interpolation error $\delta_{interp}$ is referred to work~\cite{astarita2005analysis}.

\section{Experiments}
\label{sec_4}
In this part, the performance of our diffeomorphic PIV (CDDI and FDDI) is investigated through comparison with the standard iterative PIV (FDI and CDI).  
Three representative one-pass displacement estimators are employed to form complete PIV algorithms. The estimators are cross-correlation (CC) method\footnote{https://github.com/openpiv/openpiv-python} from OpenPIV~\cite{liberzon2016openpiv} (interrogation window $32\times32$pixel$^2$, step size $8$pixel), an optical flow (OF) method\footnote{https://github.com/opencv/opencv} from OpenCV~\cite{farneback2003two,bung2017flowcv,2021opencv}, and a recent deep learning based PIV estimator (DP) from the code repository\footnote{https://github.com/erizmr/UnLiteFlowNet-PIV} of~\cite{zhang2020unsupervised}. Totally, there are 12 (3 flow estimators and 4 deformation schemes) PIV instances compared. The bicubic interpolation is used to reconstruct the warped image for all methods~\cite{astarita2005analysis,lu2021accurate}. Additional information and our implementation can be found at the project repository\footnote{https://github.com/yongleex/DiffeomorphicPIV}  for interested readers.

Due to the known ground truth, the root mean square error (RMSE) can be adopted to quantify the performance~\cite{lee2017piv,raffel2018particle}.
\begin{equation}
RMSE = \sqrt{\frac{1}{N}\mathop{\Sigma}_i |\boldsymbol{v}_{e,i}-\boldsymbol{v}_{t,i}|^2}
\end{equation}
where $\boldsymbol{v}_{e,i}=(v_{x}, v_{y})$ is the $i^{th}$ estimated vector out of $N$ points, while the $\boldsymbol{v}_{t,i}$ denotes the $i^{th}$ truth velocity. Besides the error map $|\boldsymbol{v}_{e,i}-\boldsymbol{v}_{t,i}|$ is also provided for visual assessment.



\subsection{Experiments with synthetic images}

\begin{figure}
\centering 
\subfloat[Lamb-Oseen flow]{\includegraphics[width=0.49\columnwidth]{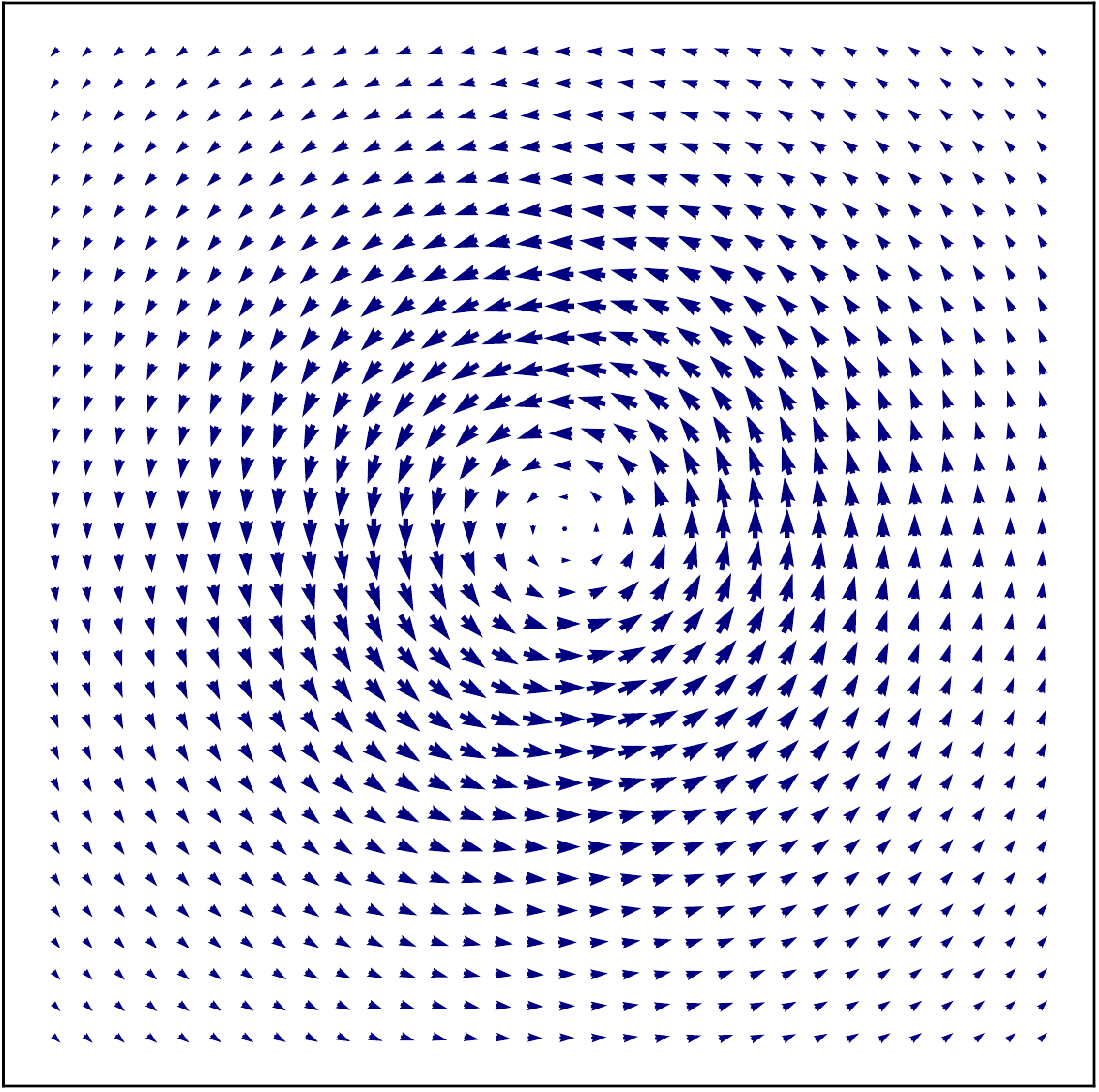}}\hspace{0.5mm}
\subfloat[Sine flow]{\includegraphics[width=0.49\columnwidth]{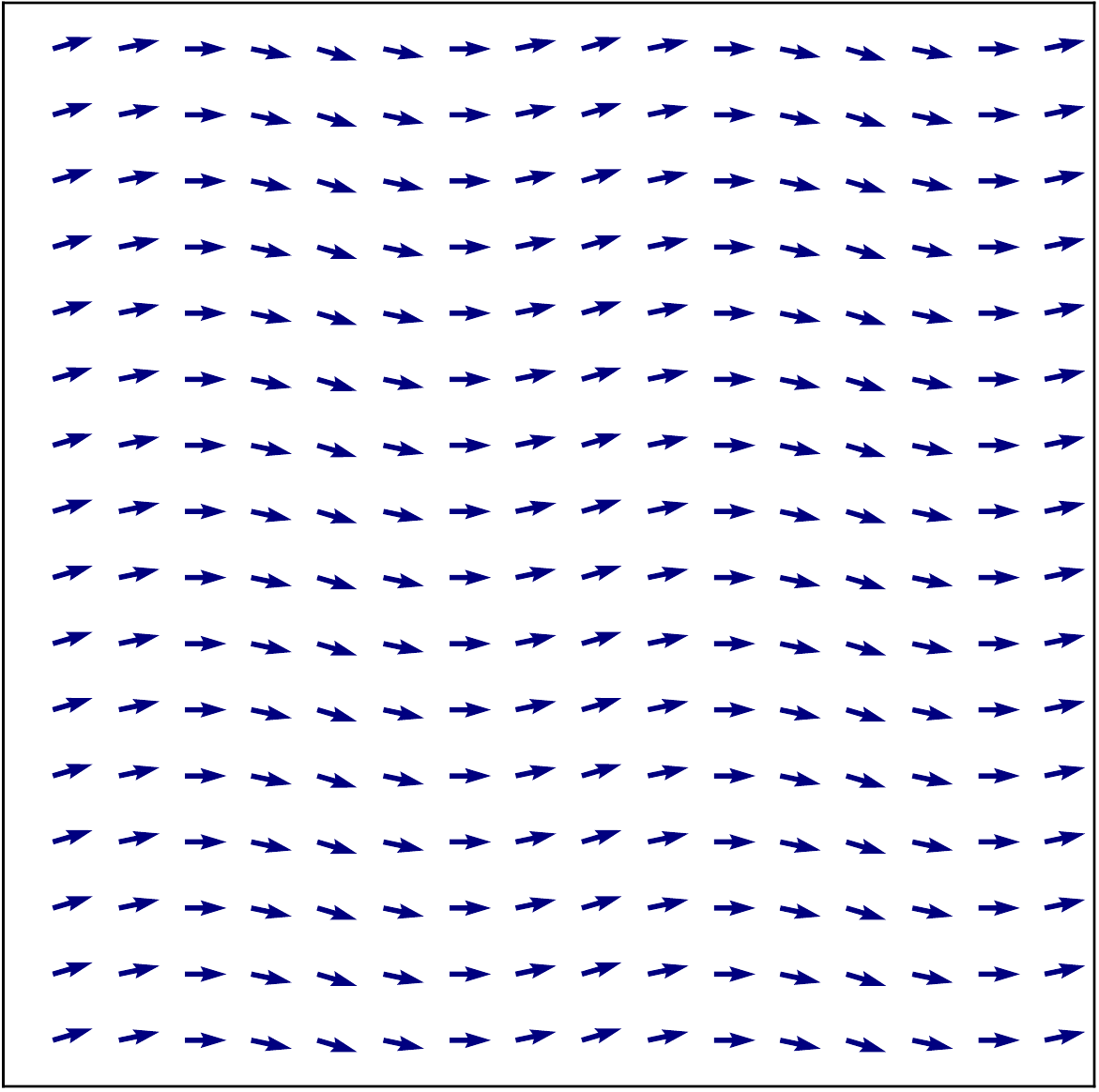}}\newline
\subfloat[A synthetic image pair]{\includegraphics[width=0.49\columnwidth]{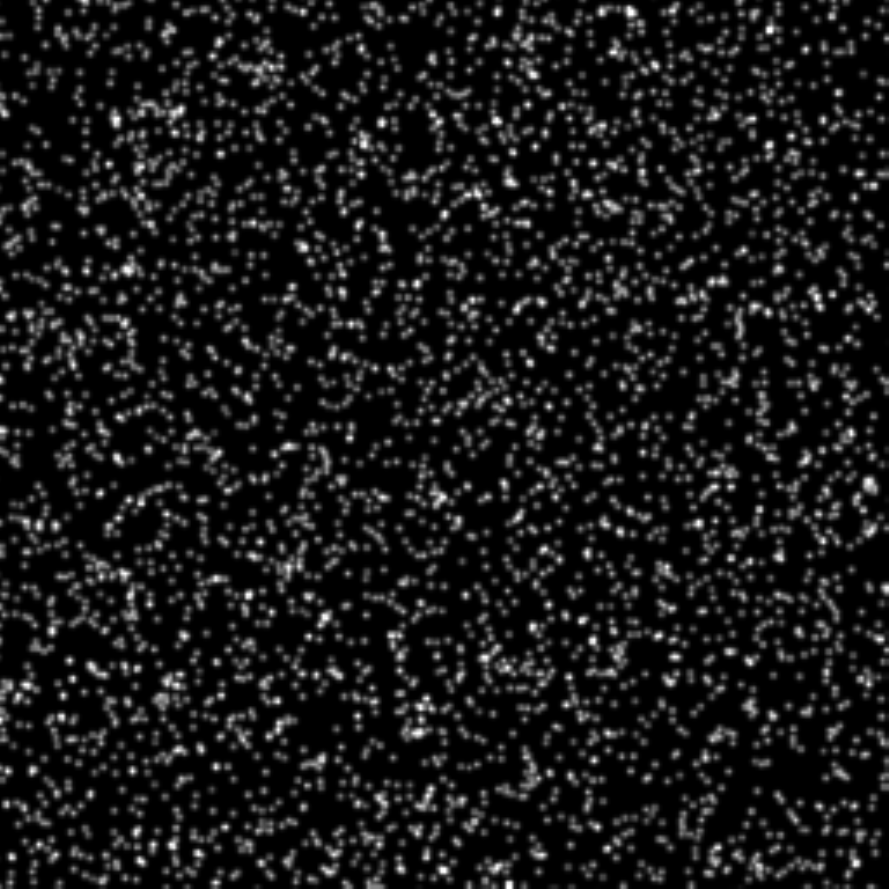}
\includegraphics[width=0.49\columnwidth]{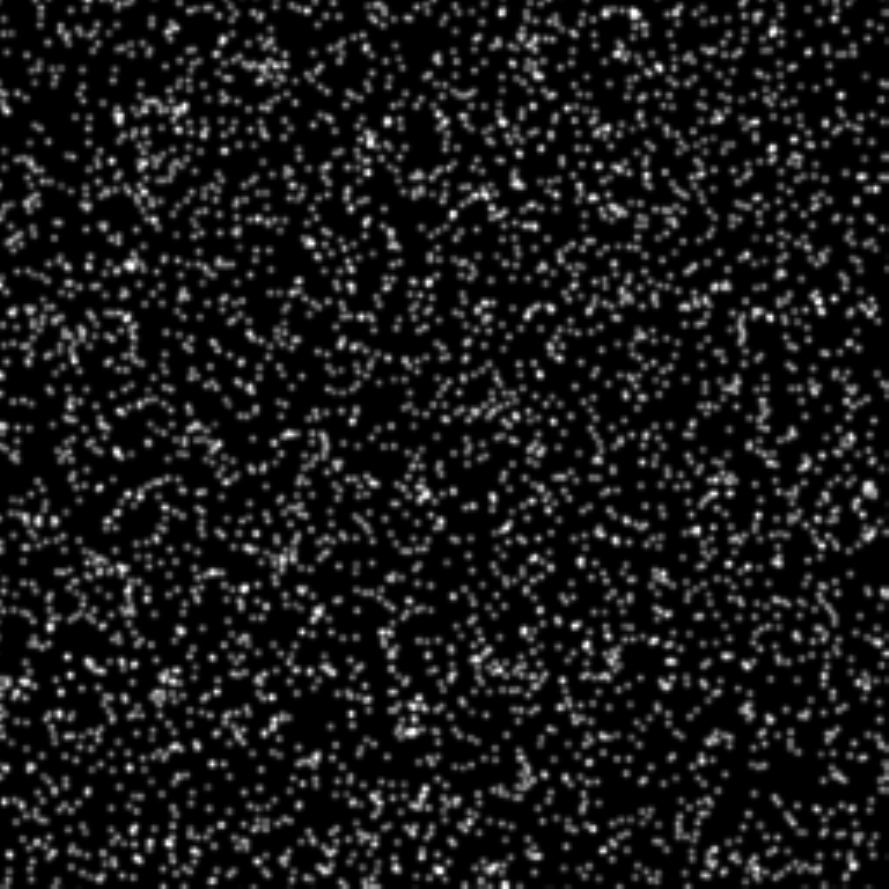}}
\caption{Two simulated flow cases with default parameters (a) and (b). A synthetic image pair for the sinusoidal flow (c)}
\label{fig_5}       
\end{figure}

One widely adopted test flow is the Lamb-Oseen vortex defined in polar coordinates~\cite{scharnowski2013effect,carlier2005second,lu2021accurate},  
\begin{equation}
\begin{split}
v_\theta & = \frac{\Gamma}{2\pi r} [1- exp(\frac{-r^2}{r_c^2})] \\
v_r & = 0 
\end{split}
\end{equation}
where $\Gamma$ is the circulation and $r_c$ denotes the vortex core radius. Another velocity field is a sine flow with sinusoidal streamlines~\cite{scharnowski2013effect}. 
\begin{equation}
\begin{split}
v_x & = \frac{c}{\sqrt{1+\frac{4\pi^2a^2}{b^2}\cos^2(\frac{2\pi x}{b})}} \\
v_y & = \frac{c\frac{2\pi a}{b}\cos(\frac{2\pi x}{b})}{\sqrt{1+\frac{4\pi^2a^2}{b^2}\cos^2(\frac{2\pi x}{b})}}
\end{split}
\end{equation}
where $(v_x, v_y)$ is the velocity in rectangular coordinate system. The $a$ and $b$ are the amplitude and the period of the sinusoidal streamline. The $c$ denotes the amplitude of the velocity. 
Under the circumstance of Lamb-Oseen flow or sine flow, the particles have curved paths instead of straight-lines.  
Fig.~\ref{fig_5} (a) and (b) displays these two flow fields with default parameters $(\Gamma=2000; r_c=40; a=6; b=128; c=5)$. 
Given a velocity field, the particle image generator (PIG) could generate the synthetic particle image pairs~\cite{raffel2018particle,lee2017piv}. 
As shown in Fig.~\ref{fig_5}(c), a noise-free image pair with size $256\times256$pixel$^2$ is synthesized with recommended PIG parameters (particle diameter $2.5$pixel, density $0.06$ppp---particles per pixel, and peak intensity $255$).

\begin{figure}
\subfloat[cross-correlation (CC) estimator]{\includegraphics[width=\columnwidth]{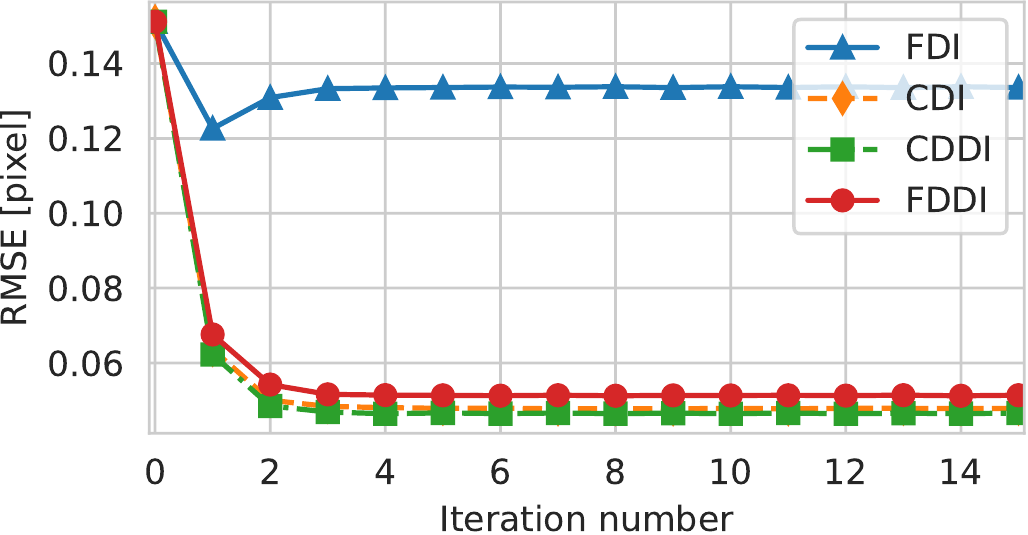}}\newline
\subfloat[optical flow (OF) estimator]{\includegraphics[width=\columnwidth]{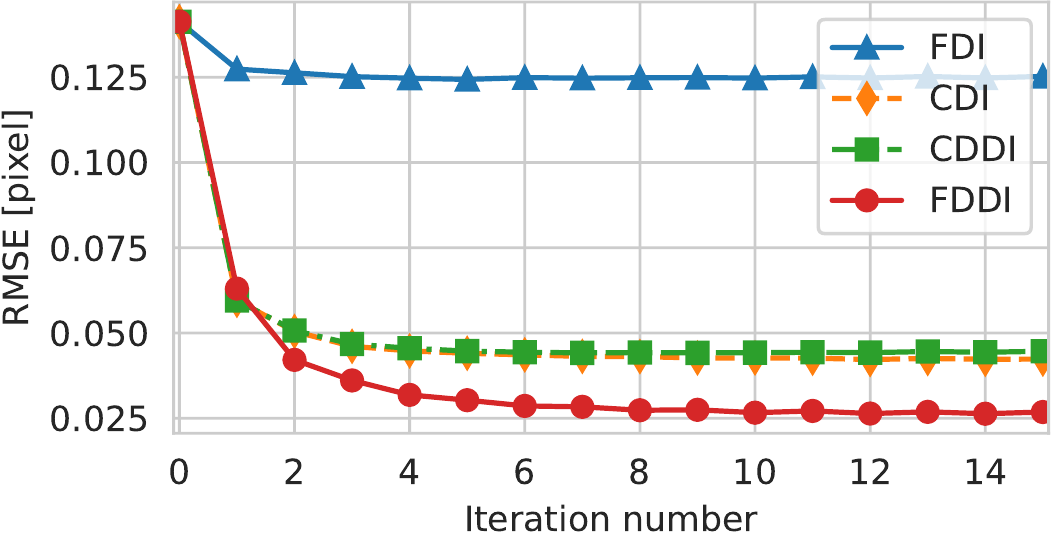}}\newline
\subfloat[deep PIV (DP) estimator]{\includegraphics[width=\columnwidth]{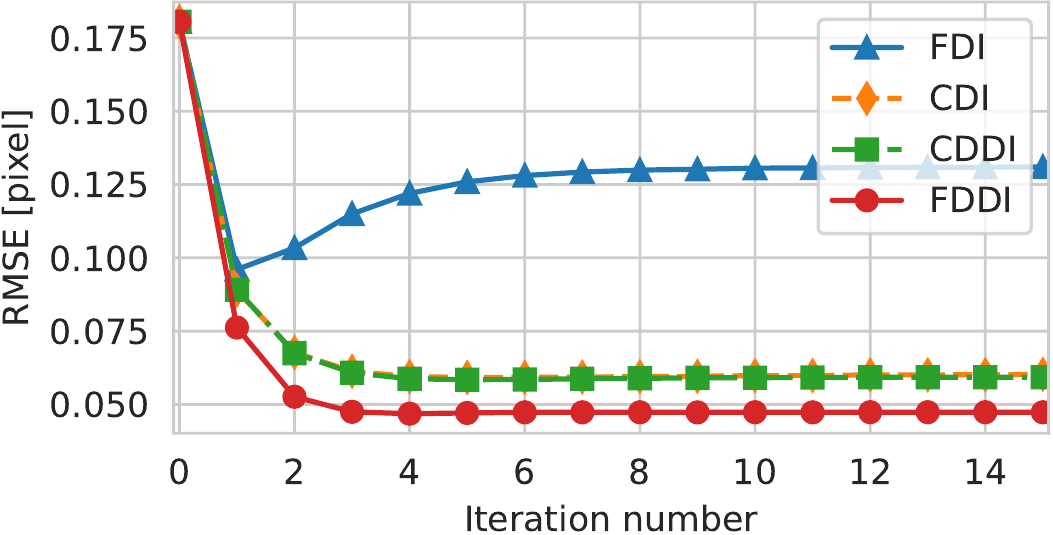}}
\caption{Convergence performance for four image deformation schemes (FDI, CDI, CDDI, FDDI) and three estimators (CC, OF, DP)}
\label{fig_6}       
\end{figure}

To investigate the convergence performance, we employ a synthetic image pair of sine flow ($c=5$). Fig.~\ref{fig_6} shows the RMSE results with a range of iterations.
For the CDI and CDDI schemes, two or three iterations are already sufficient to achieve a converged result with most of the in-plane particle image motion compensated through the image deformation~\cite{raffel2018particle}.
The FDDI achieves the convergence with three iterations for CC estimator and DP estimator.  While, the FDDI with OF estimator converges slower (with 5-7 iterations) but demonstrates more accurate result. To ensure converged results, the maximum iterations are thus set to 10 for all iterative PIV methods in the rest experiments.

\begin{table}
\caption{The performance measured by RMSE for synthetic images. The best in \textbf{bold}}
\label{tab_1}       
\centering
\begin{tabular}{|l|c|c|c|c|c|}
\hline
Case & Estimator & FDI & CDI &  CDDI & FDDI  \\
\hline
\hline
~               & CC    & $0.053$    & $0.037$    & $0.037$    & $\mathbf{0.035}$         \\  \cline{2-6}
Lamb-Oseen      & OF    & $0.048$    & $0.058$    & $0.057$    & $\mathbf{0.026}$         \\  \cline{2-6}
$(\Gamma=1000)$  & DP    & $0.064$    & $0.064$    & $0.064$    & $\mathbf{0.048}$         \\ 
\hline                  
\hline                  
~               & CC    & $0.166$    & $0.058$    & $\mathbf{0.057}$    & $\mathbf{0.057}$        \\  \cline{2-6}
Lamb-Oseen      & OF    & $0.169$    & $0.050$    & $0.049$    & $\mathbf{0.020}$        \\  \cline{2-6}
$(\Gamma=2000)$  & DP    & $0.173$    & $0.066$    & $0.065$    & $\mathbf{0.047}$        \\ 
\hline                  
\hline                  
~               & CC    & $0.368$    & $0.076$    & $0.074$    & $\mathbf{0.073}$        \\  \cline{2-6}
Lamb-Oseen      & OF    & $0.378$    & $0.038$    & $0.037$    & $\mathbf{0.022}$        \\  \cline{2-6}
$(\Gamma=3000)$  & DP    & $1.031$    & $0.592$    & $\mathbf{0.583}$    & $0.958$        \\ 
\hline                  
\hline                  
~               & CC    & $0.041$    & $0.034$    & $0.034$    & $\mathbf{0.027}$        \\  \cline{2-6}
Sin flow        & OF    & $0.037$    & $0.076$    & $0.077$    & $\mathbf{0.024}$        \\  \cline{2-6}
$(c =2.5)$      & DP    & $0.058$    & $0.093$    & $0.093$    & $\mathbf{0.049}$        \\ 
\hline                  
\hline                  
~               & CC    & $0.134$    & $0.048$    & $\mathbf{0.046}$    & $0.051$        \\  \cline{2-6}
Sin flow        & OF    & $0.125$    & $0.043$    & $0.044$    & $\mathbf{0.027}$        \\  \cline{2-6}
$(c =5.0)$      & DP    & $0.131$    & $0.060$    & $0.059$    & $\mathbf{0.047}$        \\ 
\hline                  
\hline                  
~               & CC    & $0.289$    & $0.080$    & $\mathbf{0.074}$    & $0.088$        \\  \cline{2-6}
Sin flow        & OF    & $0.276$    & $0.060$    & $0.061$    & $\mathbf{0.031}$        \\  \cline{2-6}
$(c =7.5)$      & DP    & $0.279$    & $0.053$    & $\mathbf{0.051}$    & $0.058$        \\ 
\hline
\end{tabular}
\end{table}

\begin{figure}
  \includegraphics[width=\columnwidth]{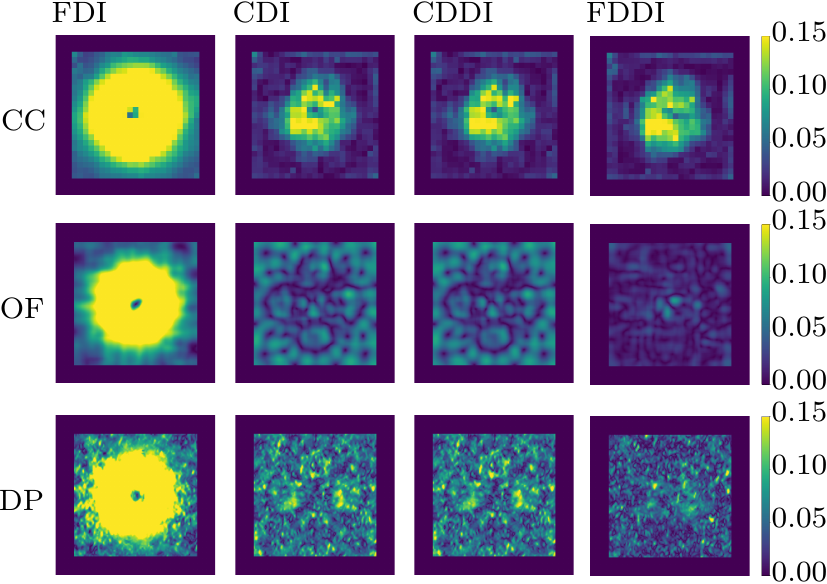}
\caption{The error distribution for Lamb-Oseen flow $(\Gamma=2000)$. The boundary effect is canceled by setting zeros values}
\label{fig_7}       
\end{figure}

\begin{figure}
  \includegraphics[width=\columnwidth]{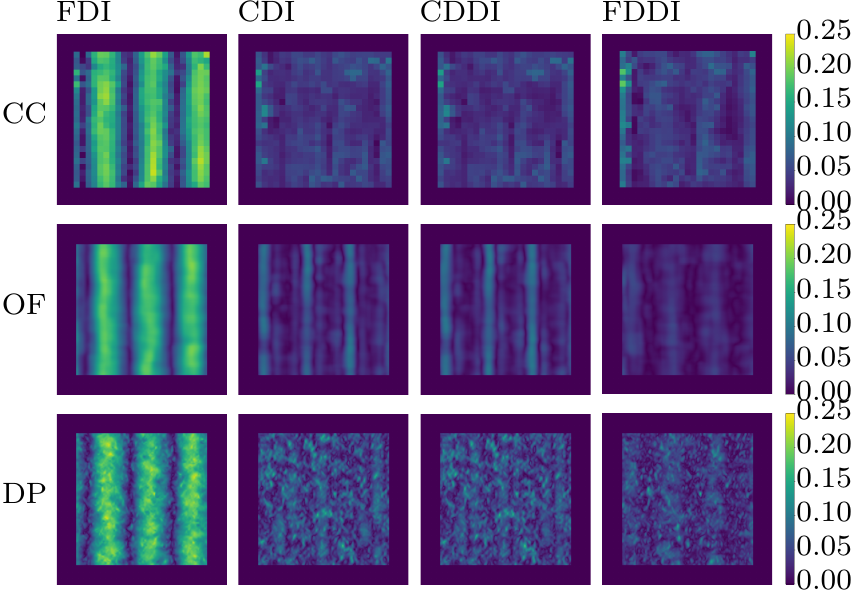}
\caption{The error distribution for sine flow $(c=5)$. The boundary effect is canceled by setting zeros values}
\label{fig_8}       
\end{figure}

To evaluate the accuracy, we adopt six simulated particle image pairs, which respectively correspond to three Lamb-Oseen flows $(\Gamma=1000;\Gamma=2000;\Gamma=3000)$ and three sine flows $(c=2.5;c=5.0;c=7.5)$. 
The performance of the different PIV instances is summarized in Table~\ref{tab_1}.
Besides, Fig.~\ref{fig_7} and Fig.~\ref{fig_8} provide corresponding error distribution of the estimations.
On all of the test flows, the FDDI with OF estimator outperforms all other combinations by a significant margin.
The poorer performance of CDDI verifies the error analysis result, that two times of interpolation of CDDI causes larger uncertainty. 
Through the comparison of different estimators, the advantages of FDDI are more obvious if the one-pass estimator could provide an accurate and dense corrector.









\subsection{Real PIV cases}

\begin{figure*}
  \includegraphics[width=2\columnwidth]{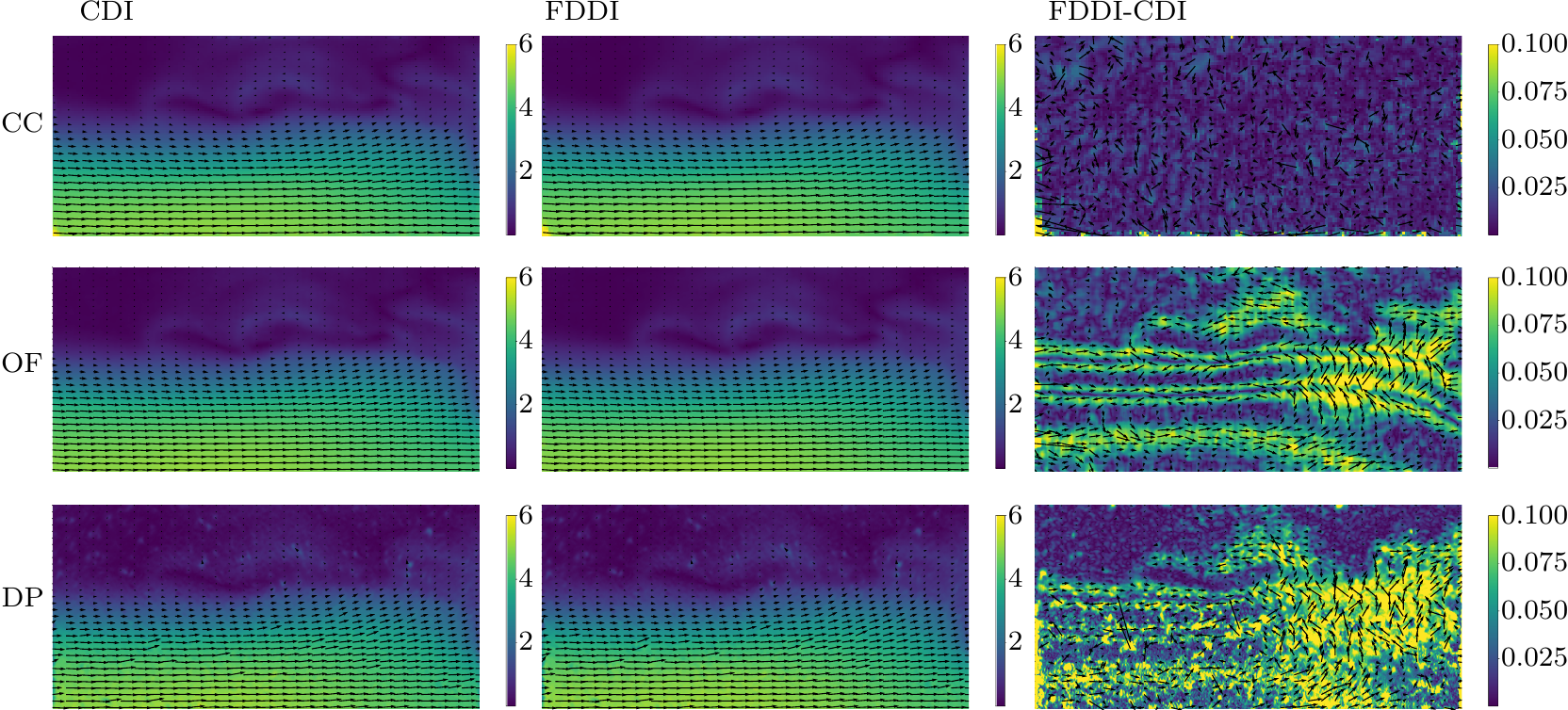}
\caption{The estimated velocity fields for a channel flow. The color background denotes the corresponding vector magnitude. The "FDDI-CDI" is short for the difference vector field between FDDI and CDI}
\label{fig_9}       
\end{figure*}

\begin{figure*}
  \includegraphics[width=2\columnwidth]{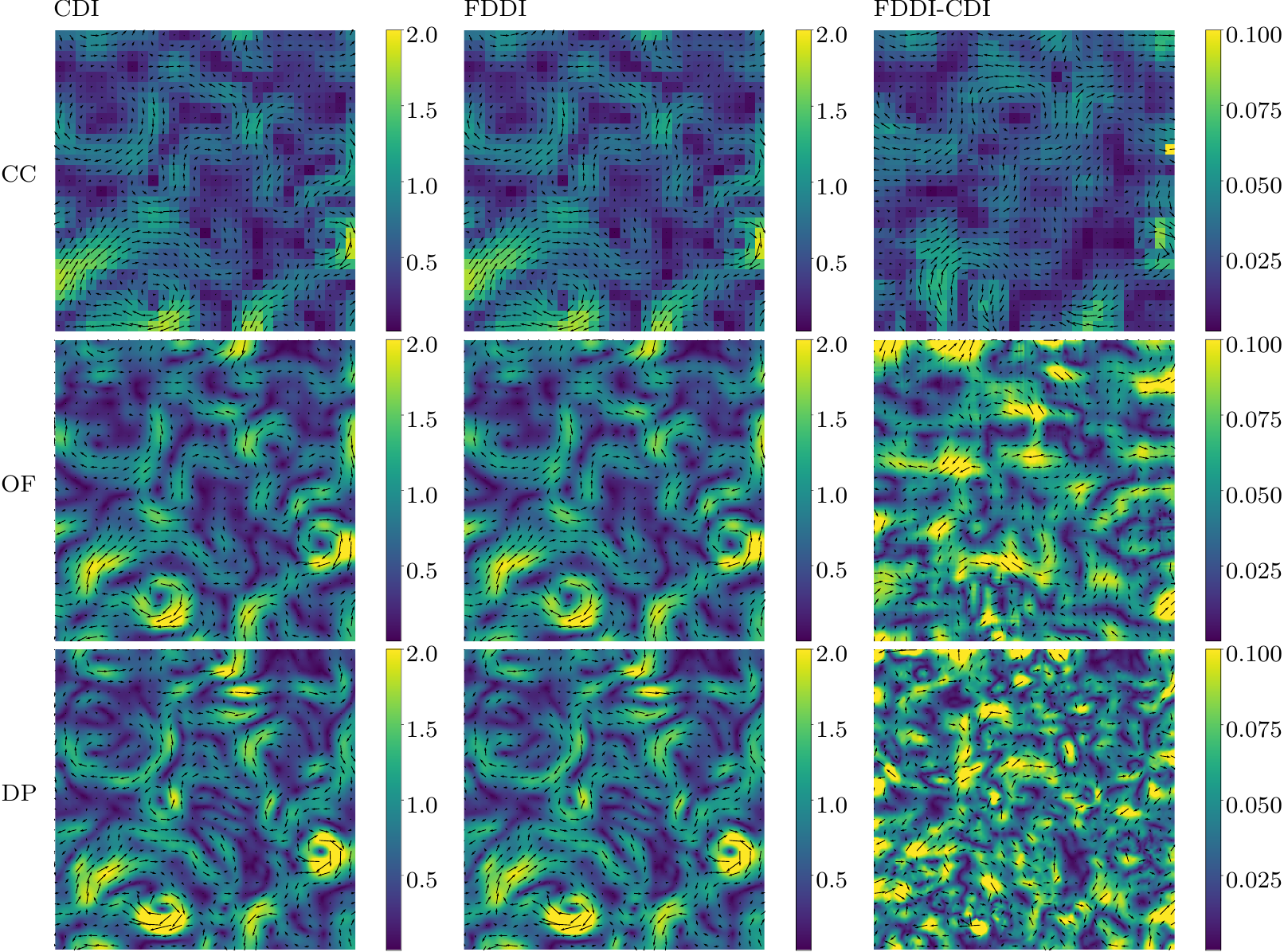}
\caption{The estimated velocity fields for a turbulent flow. The color background denotes the corresponding vector magnitude. The "FDDI-CDI" is short for the difference vector field between FDDI and CDI}
\label{fig_10}       
\end{figure*}

\begin{figure*}
  \includegraphics[width=2\columnwidth]{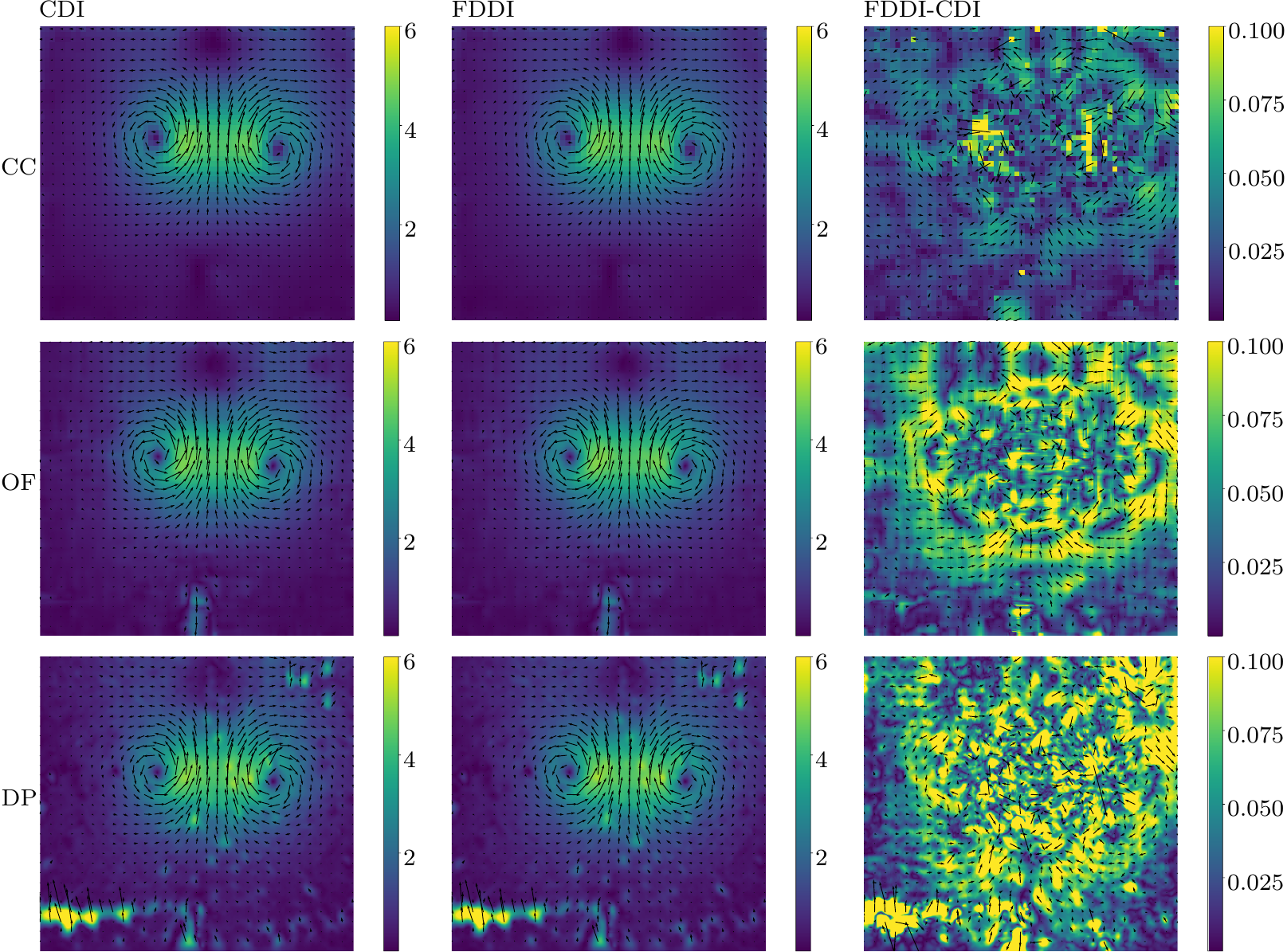}
\caption{The estimated velocity fields for a vortex pair. The color background denotes the corresponding vector magnitude. The "FDDI-CDI" is short for the difference vector field between FDDI and CDI}
\label{fig_11}       
\end{figure*}






\begin{table}
\caption{The average amplitude for difference fields (FDDI-CDI)}
\label{tab_2}       
\centering
\begin{tabular}{|l|c|c|c|}
\hline
Case & CC & OF &  DP  \\
\hline
Channel Flow    & 0.012 & 0.047 & 0.053  \\
\hline
Turbulent Flow  & 0.026 & 0.049 & 0.049 \\
\hline
Vortex Pair     & 0.033 & 0.057 & 0.068 \\
\hline
\end{tabular}
\end{table}


Three real PIV image pairs are tested in this part. The first image pair records a laminar channel flow (Case B of the third PIV challenge, images 10 and 11)~\cite{stanislas2008main,scharnowski2013effect}. The second image pair describes a 2D, homogeneous, isotropic and incompressible turbulent ﬂow (Case 1, images 1 and 2)~\cite{carlier2005second}. And the last image pair represents a vortex pair~\cite{willert1997interaction,lee2017outlier}. 
Considering the very similar performance between CDI and CDDI, we perform a meaningful comparison between the popular CDI and our proposed FDDI.


Fig.~\ref{fig_9}, \ref{fig_10} and \ref{fig_11} show the estimated velocity fields and corresponding difference vectors (FDDI-CDI). And the average magnitude of the difference vector fields are given in Table~\ref{tab_2}.
Both the CDI and the FDDI methods can predict correct velocity fields which reveal similar flow structures.
On the difference vector results,  special patterns---across different estimators---are found to be related to the flow structures. It implies the vector length of FDDI is slightly larger than that of CDI in the fast curvy streamline areas.
%
And the vector amplitude of the difference fields can reach an amazing $0.1$pixel, resulting in about $0.05$pixel of average difference amplitude. 
These values are significantly large, compared with the RMSE value of a state-of-the-art PIV estimator~\cite{lee2017piv,zhang2020unsupervised,raffel2018particle}.
Recall the shortest straight-line approximation of CDI (Fig.~\ref{fig_1}), these results imply the non-negligible systematic error of CDI in a practical PIV instrument. 
The dense estimators (OF and DP) exhibit larger average difference amplitudes.
Considering the experimental results with synthetic images, 
the curvature effect of the non-straight particle trajectory is considerably reduced with the proposed diffeomorphic PIV (a combination of FDDI and accurate dense estimator).

\section{Conclusion}
\label{sec_5}
A diffeomorphic PIV technique is proposed to reduce the curvature effect of the non-straight particle trajectory. 
Considering the significant difference between an instantaneous velocity field and the cumulative deformation vector field,
our diffeomorphic PIV is devoted to estimating the exact velocity vector field instead of computing the deformation field like other existing PIV techniques.
Two diffeomorphic deformation schemes (CDDI and FDDI) are implemented for diffeomorphic PIV as results of iterative optimization. 
Both CDDI and FDDI move the particles (image) along the streamline of an estimated velocity field. 
Tested on synthetic images, the combination of FDDI and an accurate dense estimator (e.g., optical flow) achieves a significant accuracy improvement---cutting the error by half--- for complex flow measurement. On real PIV images, the FDDI provides a larger estimation which is more reasonable due to the curved path. Moving forward, we plan to extend our diffeomorphic PIV beyond classical planar PIV to other applications including Tomo-PIV, 3D-PIV, and TR-PIV.

\appendix[Proof of the Variation Approximation]
The variation of the deformation function is
\begin{equation}
\begin{split}
\Delta \boldsymbol{\phi} &= \boldsymbol{\phi_{\boldsymbol{v}_k+\Delta {\boldsymbol{v}}}} - \boldsymbol{\phi}_{\boldsymbol{v}_k}\\
&=\boldsymbol{\psi^{(1)}_{\boldsymbol{v}_k+\Delta {\boldsymbol{v}}}} - \boldsymbol{\psi}^{(1)}_{\boldsymbol{v}_k}
\end{split}
\end{equation}
Considering the variation of $\boldsymbol{\psi}^{(t)}$,
\begin{equation}
\begin{split}
\Delta \boldsymbol{\psi}^{(t)} &=\boldsymbol{\psi^{(t)}_{\boldsymbol{v}_k+\Delta {\boldsymbol{v}}}} - \boldsymbol{\psi}^{(t)}_{\boldsymbol{v}_k}
\end{split}
\end{equation}
Obviously, $\Delta \boldsymbol{\phi} =\Delta \boldsymbol{\psi}^{(1)}$. According to the transport equation, we have
\begin{equation}
\begin{split}
\frac{\partial \boldsymbol{\psi}_{\boldsymbol{v}_k}^{(t)}}{\partial t} &= \boldsymbol{v}_k(\boldsymbol{\psi}_{\boldsymbol{v}_k}^{(t)})\\
\frac{\partial \boldsymbol{\psi}_{\boldsymbol{v}_k+\Delta {\boldsymbol{v}}}^{(t)}}{\partial t} &= [\boldsymbol{v}_k+\Delta {\boldsymbol{v}}](\boldsymbol{\psi}_{\boldsymbol{v}_k+\Delta {\boldsymbol{v}}}^{(t)})\\
\end{split}
\end{equation}
That says, 
\begin{equation}
\begin{split}
\frac{\partial \boldsymbol{\psi}_{\boldsymbol{v}_k}^{(t)}}{\partial t} +\frac{\partial \Delta \boldsymbol{\psi}^{(t)}}{\partial t} &= \boldsymbol{v}_k(\boldsymbol{\psi}_{\boldsymbol{v}_k}^{(t)}+\Delta \boldsymbol{\psi}^{(t)})+\Delta {\boldsymbol{v}}(\boldsymbol{\psi}_{\boldsymbol{v}_k}^{(t)}+\Delta \boldsymbol{\psi}^{(t)})\\
\end{split}
\end{equation}
Considering the first transport equation, we have
\begin{equation}
\label{eq_20}
\begin{split}
\frac{\partial \Delta \boldsymbol{\psi}^{(t)}}{\partial t} &= \boldsymbol{v}_k(\boldsymbol{\psi}_{\boldsymbol{v}_k}^{(t)}+\Delta \boldsymbol{\psi}^{(t)})-\boldsymbol{v}_k(\boldsymbol{\psi}_{\boldsymbol{v}_k}^{(t)})+\Delta {\boldsymbol{v}}(\boldsymbol{\psi}_{\boldsymbol{v}_k}^{(t)}+\Delta \boldsymbol{\psi}^{(t)})\\
\end{split}
\end{equation}
Because the variations of the functions $\boldsymbol{v}$ and $\boldsymbol{\psi}$ are small, and the velocity and corresponding variation are assumed smooth.
\begin{equation}
\begin{split}
\boldsymbol{v}_k(\boldsymbol{\psi}_{\boldsymbol{v}_k}^{(t)}+\Delta \boldsymbol{\psi}^{(t)})\approx \boldsymbol{v}_k(\boldsymbol{\psi}_{\boldsymbol{v}_k}^{(t)})\\
\Delta {\boldsymbol{v}}(\boldsymbol{\psi}_{\boldsymbol{v}_k}^{(t)}+\Delta \boldsymbol{\psi}^{(t)}) \approx\Delta {\boldsymbol{v}}(\boldsymbol{x}) 
\end{split}
\end{equation}
Hence, the Eq.~(\ref{eq_20}) becomes to 
\begin{equation}
\begin{split}
\frac{\partial \Delta \boldsymbol{\psi}^{(t)}}{\partial t} &\approx \Delta {\boldsymbol{v}}(\boldsymbol{x})\\
\end{split}
\end{equation}
To this end, the conclusion is arrived with initial $\Delta \boldsymbol{\psi}^{(0)}=0$, 
\begin{equation}
\begin{split}
\Delta \boldsymbol{\phi} =\Delta \boldsymbol{\psi}^{(1)}\approx\Delta {\boldsymbol{v}}
\end{split}
\end{equation}
Note that this is an unbiased approximation, i.e., the $\Delta \boldsymbol{\phi}=\Delta \boldsymbol{v}$, if the smooth variation of the velocity field $\Delta \boldsymbol{v}$ is closed to zero.

\section*{Acknowledgment}
We would like to thank all the professional editor
and reviewers for the substantial effort and expertise that contribute to
this work. This work was supported by the National Natural
Science Foundation of China under Grant No. 51905502.

\ifCLASSOPTIONcaptionsoff
  \newpage
\fi

\begin{IEEEbiography}[{\includegraphics[width=1in,height=1.25in,clip,keepaspectratio]{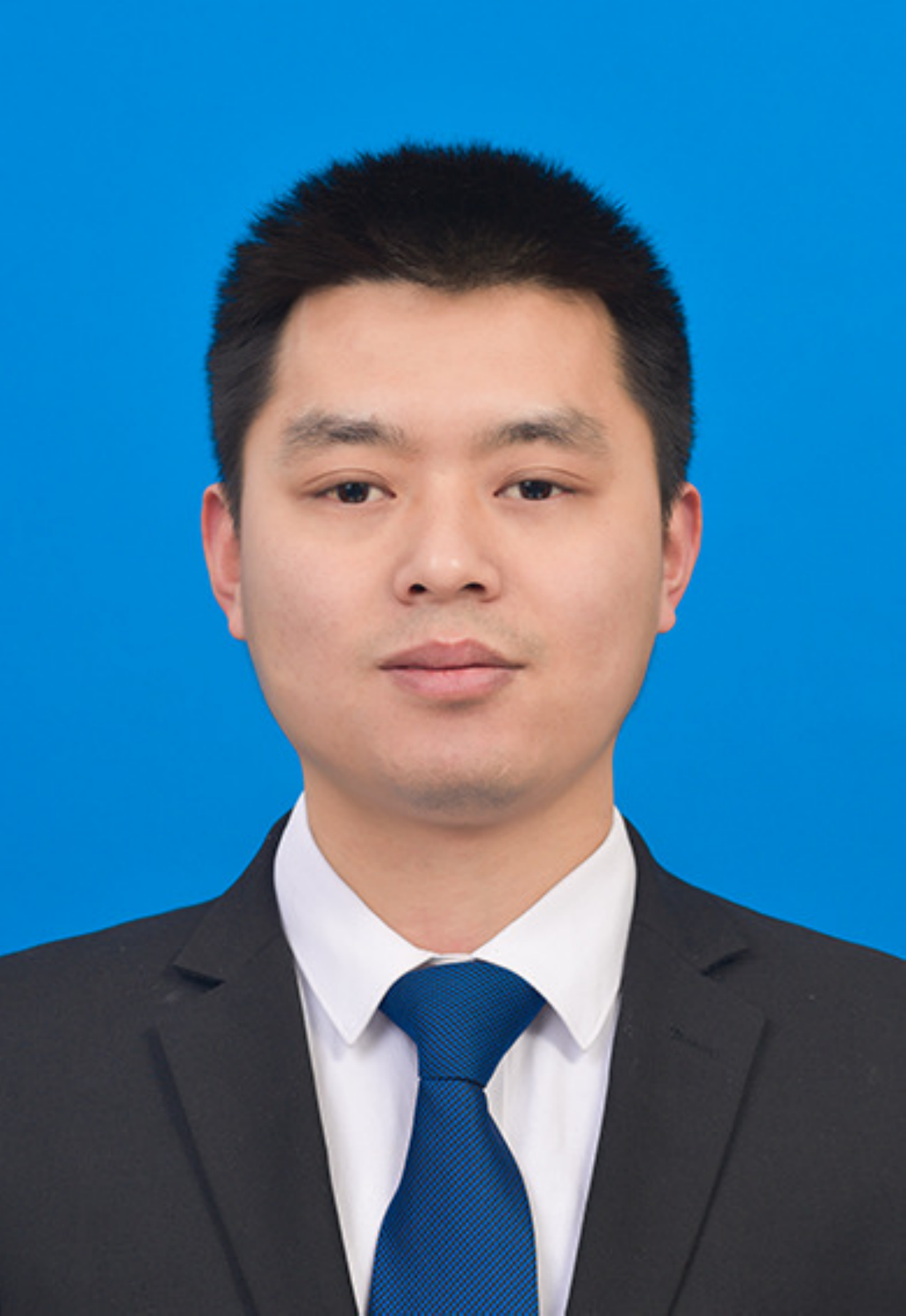}}]{Yong Lee}
received his BSc and Ph.D. degrees from Huazhong University of Science and Technology (HUST), Wuhan, P. R. China, in 2012 and 2018 respectively. He worked at Cobot (Wuhan Cobot Technology Co., Ltd.) as an algorithm scientist from 2018 to 2019.  He was a Research Fellow with Department of Computer Science at National University of Singapore. His research interests include image processing, particle image velocimetry, machine learning and deep learning.
\end{IEEEbiography}

\begin{IEEEbiography}[{\includegraphics[width=1in,height=1.25in,clip,keepaspectratio]{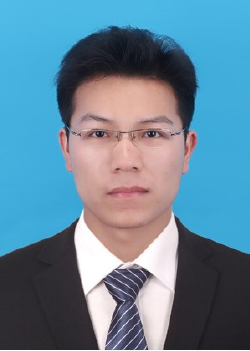}}]{Shuang}\textbf{Mei}, an associate professor of the China University of Geoscience since 2018, outstanding young talents of 2020 Huanghe Talents Plan in WuHan. He recieved the B.S. degree in Mechanical Electronic and Information Engineering from China University of Geoscience, Wuhan, China, in 2012, and the Ph.D degrees in School of Mechanical Science and Engineering from the Huazhong University of Science and Technology, Wuhan, China, in 2017.  His current research interests include machine vision, deep learning technology and its application (target recognition and location, anomaly detection and digital image correlation).
\end{IEEEbiography}







\end{document}